\documentclass[review]{elsarticle}
\usepackage{graphicx}
\usepackage{graphicx}
\usepackage{color}
\usepackage{lineno,hyperref}
\usepackage{import}
\usepackage{algorithmic}
\usepackage{algorithm}
\usepackage{amsmath,amssymb,amsfonts}

\begin{document}

\begin{frontmatter}

\title{Visual Place Representation and Recognition from Depth Images}

\author{Farah Ibelaiden, Slimane Larabi}

\address{Computer Science Department, USTHB University\\
RIIMA Laboratory, 16111, Algeria}

\begin{abstract}
This work proposes a new method for place recognition based on the scene architecture. From depth video, we compute the 3D model and we derive and describe geometrically the 2D map from which the scene descriptor is deduced to constitute the core of the proposed algorithm. The obtained results show the efficiency and the robustness of the propounded descriptor to scene appearance changes and light variations.
\end{abstract}


\end{frontmatter}


\section{Introduction}\label{sec:introduction}

Visual localization has attracted researchers efforts because it constitutes one of the important challenges for auto-positioning of visually impaired, augmented reality and intelligent robotics \cite{liang2013image} \cite{acharya2019bim}.
Place recognition methods for outdoor achieve high accuracy; however; they are useless indoors. Thus, the place recognition in indoor scenes where people spend approximately 90\% of their time \cite{diffey2011overview} remains a challenge.

The proposed methods may be differentiated according to their objectives into systems that recognize previously visited places and systems that deliver the position and orientation ($6DOF$) of the visual acquisition device. \\
The permanent changes of scenes induce the scene descriptors modification and lead to a state of the art localization system failure \cite{lowry2015visual}. That's why the scene descriptors dataset needs to be updated in order to consider the scene modifications that's the factor which was unkempt-ed by state of the art methods which have focused only on betterments of positioning approaches.
In this insight, we aim through our work to propose
a place recognition system based on architectural features to recognize labeled places 
 whose dataset is weight-light and does not need updates even in case 
 where the scene undergo major modifications.

Section \ref{sec:stateoftheart} is devoted to related works. in section \ref{sect:scenedesc}, we present our method for scene description from a sequence of depth images. In section \ref{visual}, we introduce the dataset construction framework and the query scene identification method. 
Experiments are detailed in section \ref{exp} and finally the paper is concluded in section \ref{conclusion}.

\section{Related works}\label{sec:stateoftheart}

The visual localization methods may be categorized into:\\
\textbf{Feature based methods:} Many kind of features have been used in order to tackle two main problems: excerpt the steady features over time, render these features insensitive to variation of illumination and viewing angle. These methods can broadly be splitted  into 2D based methods, 3D based methods and topological based methods.\\

The 2D based methods have tackled the problem of visual localization as an image retrieval problem  \cite{sahdev2016indoor}. \cite{knopp2010avoiding} addresses one of the key problems in place recognition that is the presence of commons objects to different places (trees, road markings...); and demonstrates that the localization can be significantly improved by automatic identification and suppression of these objects from database images Whereas, authors of \cite{torii2013visual} have proved that  these commons structures can form distinguishing features for many places by a modification of their weights in the bag of visual word model.
\\
The 3D based methods require 3 dimensional model of the scene, a large dataset of features extracted from database images used for 3D reconstruction \cite{feng2017visual} \cite{chen2017vision} and efficient retrieval method to search the most similar dataset image to the query in order to calculate its location (6DOF pose) \cite{gao2019mobile} \cite{mur2015orb} \cite{mur2017orb}. We distinguish from them the  2D-3D based methods as in \cite{deretey2015visual} where an approach has been proposed to compute the 6DOF position of a camera using  $PNP$ algorithm \cite{lepetit2009epnp}, the results have demonstrated the precision of computed position. Whereas, In \cite{feng2017visual} the 6DOF position of the camera has been computed on the basis of epipolar constraint, the efficiency of the proposed solution decreases as the area of indoor scene increases due to accumulated alignment errors.
\\
The 3D-2D based methods match features of 3D model against features of query image.
In \cite{sattler2017efficient} authors have compared between 2D-3D based methods and 3D-2D based methods and affirm that 3D-2D search reach a better effectiveness but it is slower than 2D-3D search notably for large scale scenes because model features are more than query features. Based on this insight, authors in  \cite{li2010location} attempted  to prove that 3D-2D matching can be  more quickly than  2D-3D  through their modified 3D-2D based matching approach however results showed significant reduction in term of  precision.\\
The topological based methods are founded on adjacency (topological) maps that depict the environments by nodes representing locations to which a set of reference images are assigned and arcs for the adjacency relationships between these locations.  Like in \cite{ulrich2000appearance} where a real time system was proposed to track the mobile robot’s position by limiting the search of the most similar reference image to the query image on the currently believed location and its immediate neighbors identified from the topological map to reduce the localization time.
\\

\textbf{The neural network based methods} use features extracted from a network as an image representation \cite{he2015spatial}. They may be decomposed into: (1) methods that tackle the problem of place recognition as classification problem \cite{oertel2020augmenting} as in \cite{fan2020visual} where authors propounded enhancement on the retrieval stage of the visual localization system \cite{irschara2009structure} by a combination of multiple learning-based feature extractors. (2) methods that deal the problem of localization as regression of pose estimation. Such as in \cite{kendall2015posenet} where researchers have proposed a simple algorithm that consists of a convolutional neural network trained end to end to regress the camera orientation and position in real time.\\

\textbf{Segmentation based methods} use results of object segmentation as basis information for  place recognition (for example if there is a bed it is a bedroom) \cite{deng2017unsupervised} \cite{finman2015toward}.\\

\textbf{Architecture based methods} appeared recently, founded on the scene architecture.
In \cite{boniardi2019robot}, a robot positioning system has been proposed using a CNN to  predict edges of image and compute robot position by matching the extracted edges to the  environment floor-plan.

In-depth study of the state of the art methods reveals that despite all the improvements made in feature-based methods, they remain highly sensitive to high light variations and large modifications of decor that are frequent in daily life.
Learning based methods require a large amount of data for training phase,  which is costly in terms of computation time. Furthermore, their  generalization ability in large dynamic indoor environments is low. Movable objects (such as chairs) induce the accuracy decrease of segmentation based place recognition methods because they do not meet the criterion of stability over time. latterly, lot of interest has been attracted towards architectural features  because they are unchangeable and stable.
Most of the proposed methods require data-set updates due to the standing local modifications of the indoor environment.

In our previous work \cite {ibelaiden2020scene}, we proposed an architecture-based scene descriptor calculated from partial depth views (captured by surrounding kinect). Plans were identified and aligned in order to build the 3D scene model, walls were located and projected to define the 2D map. Which is described geometrically to define the scene descriptor. As, this descriptor didn't reflect the complete architectural features of the captured scene, we therefore focus through this work on improving it by integrating information of doors, windows, and stairs.

Our aim through this work is:
\begin{itemize}
    \item To propose from depth video an improved version of the architecture based scene descriptor previously proposed in  \cite{ibelaiden2020scene} by including information of windows, doors, and stairs on the basis of simple architectural norm  in order to increase the descriptor discrimination.
      \item To propound a Place Recognition method based on non-rearrangeable parts of the scene (walls, stairs) that represent stable features over time, independent on scenery changes. That makes our method able to recognize places whose descriptor is inserted in the database even years after its construction despite the standing local changes which may occur in the indoor environment.
   \end{itemize}

\section{Scene description from depth images}\label{sect:scenedesc}
\begin{figure}[h!]
	\centering
	\includegraphics[width=10cm]{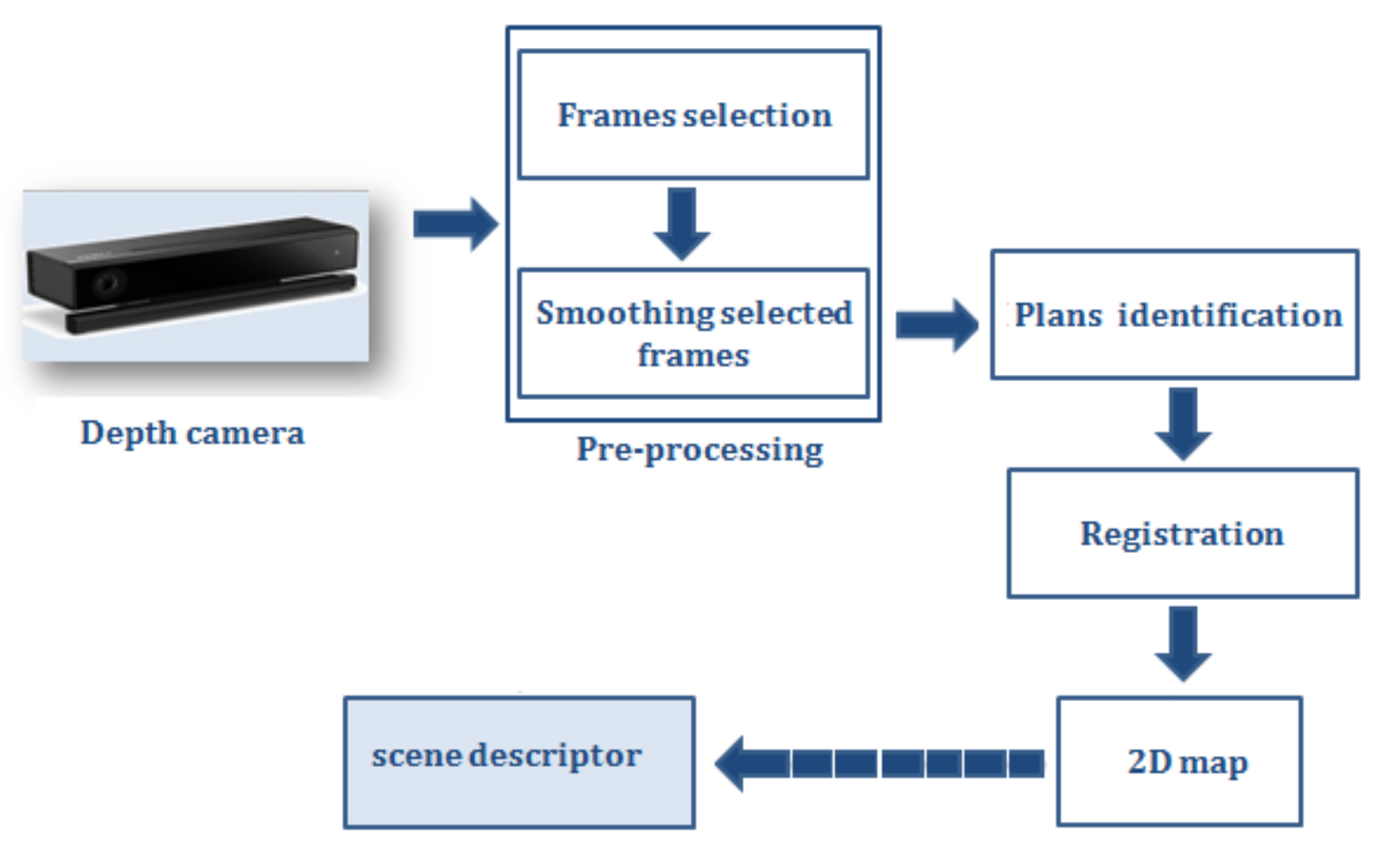}
	\caption{Scene description process.}
     \label{scene_desc_g}
\end{figure}

From a sequence of depth images of an indoor environment, the scene descriptor is computed following the diagram of figure \ref{scene_desc_g}.

\subsection{Plans extraction}

That starts by a pre-processing stage which includes (1) the frames selection that consists on selecting a key-frame after each constant number of frames \cite{williams2008image} \cite{williams2009comparison}. This implies a decrease of time processing and data storage space. (2) Smoothing key-frames by applying the median filter.

\begin{figure}
	\centering
	\includegraphics[width=12cm]{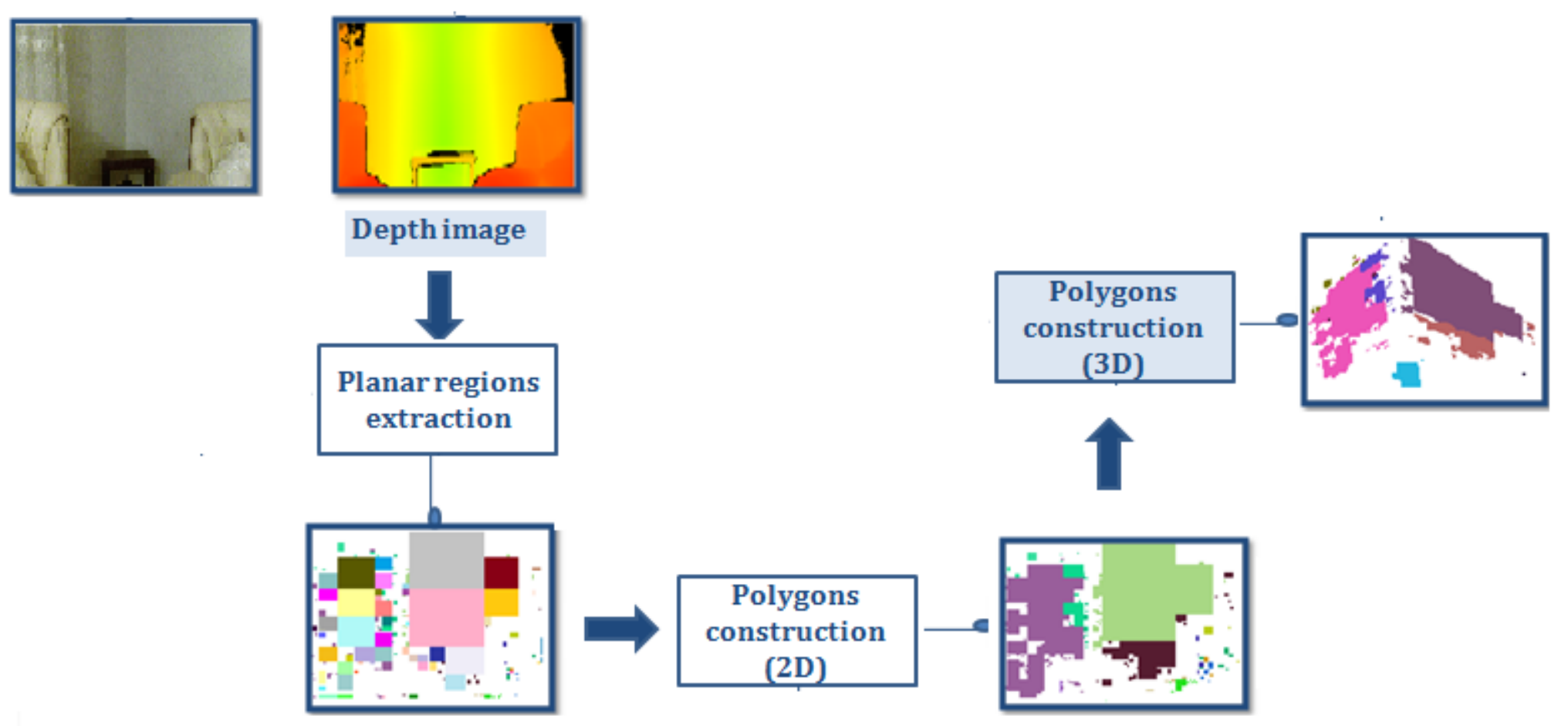}
	\caption{Plans identification from key frames. From depth image (RGB images is not used in the framework) and the steps followed for plans extraction.} \label{fig:process}
\end{figure}

 Then, the planar regions are extracted from each selected key frame (see figure \ref{fig:process}) by dividing recursively the depth image  using the quad tree algorithm \cite{hunter1979operations} into rectangular areas (defined by their left-top pixel, their width, their height and the ensemble of points that compose them). The intermediate regions $R$ that satisfy to both conditions of flatness $F_l$ \cite{xing2018extracting} and smoothness $S_m$ are considered as planar whereas the others that are not small enough (whose width is greater than a fixed threshold) are recursively subdivided into four regions.
The flatness $F_l$ of an intermediate region is verified by the equation \ref{eq:1}:

\begin{equation}
\label{eq:1}
F_l(R) =
\left\{
	\begin{array}{l l}
		1 &\left\{
			\begin{array}{l l}
				& MSE(R) < T_m \ and \\
				& C_r(R) < T_c
			\end{array}
		  \right.,\\
		& \space\\
		0 & otherwise\\
	\end{array}
\right.
\end{equation}

Where: $MSE(R)$ represents the value of Mean Square Error computed using the method proposed in \cite{holzer2012adaptive}, the $C_r(R)$ is the curvature of a region in the depth image.

So if the $MSE$ and the curvature are lower than $T_m$ and $T_c$ (depicting respectively the MSE and Curvature thresholds experimentally fixed to $1.5$ $  e^{-7}$ and $0.0008$), the region R is considered as flat.
\\\\
And to ascertain the smoothness of a region, the depth change indication $(DCI)$ map \cite{Holzer} is firstly computed in order to detect the wide changes of depth noted $I_V$, defined formally by the equation \ref{eq:2}:

\begin{equation}
\label{eq:2}
DCI(u, v) = \left\{
\begin{array}{l l}
1 & \max\limits_{(m,n)\in F} {\left| {I_V(u, v)} - {I_V(m, n)}\right|} \leq {f_s}, \\
0 & otherwise\\ \end{array} \right.
\end{equation}
where $ F$ is the $4-$neighbors of $(u,v)$, $f_s$ is the smoothness threshold  function,  $(m,n)$ is the coordinates of a pixel in the depth image. \\
Once the $DCI$ is calculated, the region smoothness $S_m(R)$ may be verified by:

\begin{equation}
\label{eq:3}
S_m(R) = \left\{
\begin{array}{l l}
1 &  \space \space |R| = \sum\limits_{(u,v)\in R} DCI(u,v),\\
0 & otherwise  \\ \end{array} \right.
\end{equation}

$|R|$ is the size of the intermediate region $R$.

As interpreted by the equation \ref{eq:3} a region is considered as smooth if their pixels have not a high differences of depth with their neighbors.

\subsection{Planar regions clustering}

For each key frame, the extracted planar regions are clustered into distinct polygons. Each polygon regroups those that belong to the same plan. Two regions $R1$ and $R2$  are considered as belonging to the same plan if the difference between their normal vectors and the distance between them (distance between their centers) are small enough:
\begin{equation}
\label{eq:4}
 \left\{
\begin{array}{l l}
				& || \vec{n1} -  \vec{n2} || < L_{nom}  \\
				& | d_1 -  d_2 | <  L_{dist}
\end{array} \right.
\end{equation}

Where $\{ \vec{n_1}(a, b, c),  d_1\}$ and $\{\vec{n_2}(e, f, g),  d_2\}$ are parameters of planar regions $R1$ and $R2$  characterized by normal vectors $n_1$ and $n_2$, distances from regions to the coordinate system center $d1$, $d2$.
\\
The clustering is done for the three main reasons:
(1) It reduces the data storage space because the space needed to store polygons (depicted by a set of vertices) is significantly lower than the space required for planar regions (defined by all points contained in their surfaces). (2) It contributes to decrease the processing time.
(3) The existence of effectual libraries for polygons manipulation such as clipper \cite{vatti1992generic}.

\subsection{3D and 2D map computation}

In this step, we apply a registration of key frames in order to compute the 3D map. As the global coordinate system (of the scene) is attached to the first key frame, we estimate the geometric local transformations  $ T_i, i \in {1..n}$ (where n represents number of key frames) between each two consecutive depth key frames $(I_i, I_{i + 1})$, and the global transformation $ \Delta T_n$ for a key frame $I_n$, is obtained as the product of all previous  transformations $\Delta T_i$ (temporally ordered), defined by:

\begin{equation}
\label{product}
\Delta T_n = \prod\limits_{i=1}^{n} \Delta T_i
\end{equation}

\begin{figure}[h!]
	\centering
	\includegraphics[width=10cm]{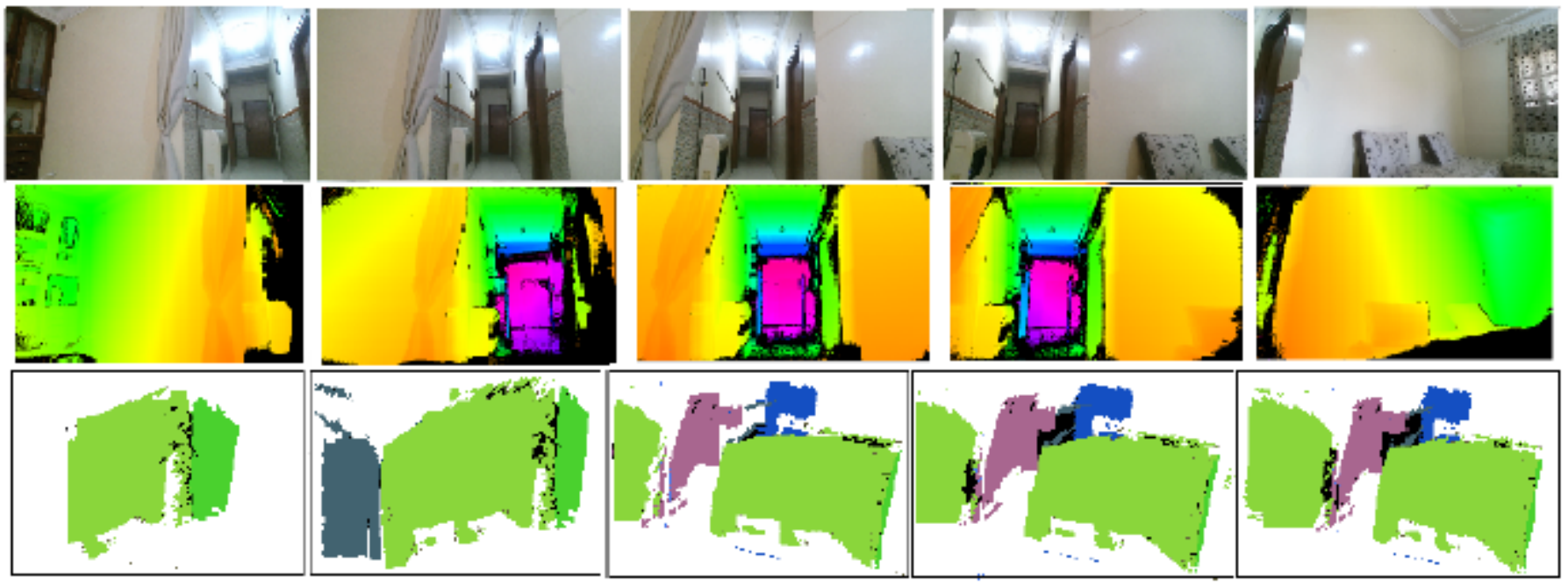}
	\caption{From top to bottom: RGB images acquired by rotating a Kinect sensor, The associated depth frames, the results of the 3D map construction.}
     \label{fig3}
\end{figure}

The set of polygons of the first depth key frame do not undergo any transformation while the set of polygons extracted from depth key frames $I_i, i \in {2.. n}$, will be transformed by the transformation $\Delta T_i$ in order to be merged with polygons of the previous frames as illustrated by figure \ref{fig3}.

Once the ground is located using the geometrical method presented in \cite{zatout2019ego} \cite{zatout2021}, the 2D map is defined as the orthogonal projection of the 3D map on the ground by taking into account the architectural norms of  doors and windows positioning following the international organization for standardization (ISO): position of windows between 1.2 meter (m) and 2.10 m, door height set to 2.10 m. We specify that extended windows to ground are also considered as doors.

the 2D map derivation from 3D model is made as follow:\\

\begin{figure}[h!]
	\centering
	\includegraphics[width=8cm]{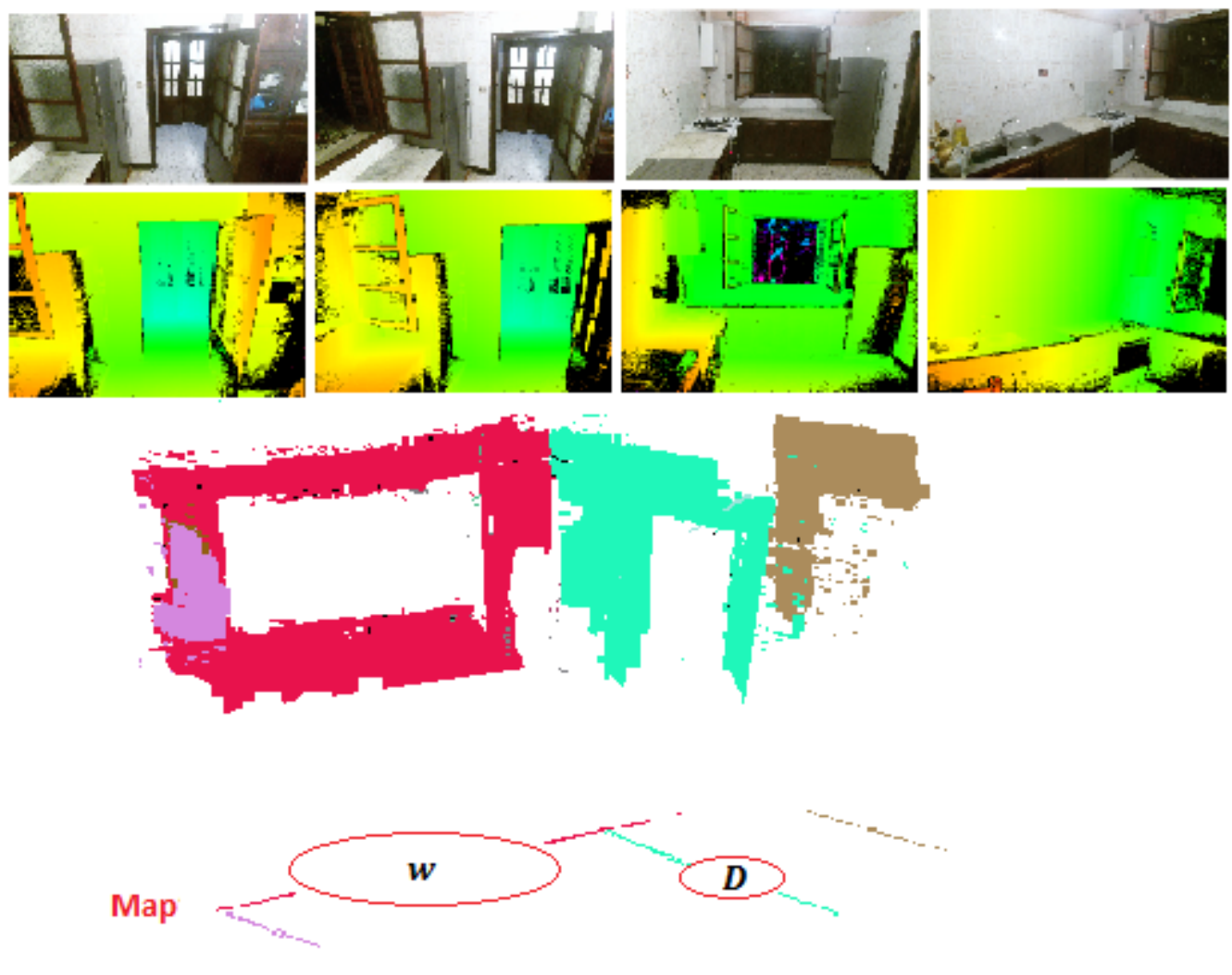}
	\caption{Example illustrating the overtures of window and door computed in the indoor scene shown by the RGB images in the first row and depth images in the second row. The door overture and the window aperture are spotted in the 2D map.}
     \label{fig4}
\end{figure}

-The areas of the orthogonal polygons to the ground \cite{ibelaiden2020scene} are computed using detailed method in \cite{vatti1992generic} then these polygons are stored in an auto-balanced red-black tree \cite{hinze1999constructing} in descending order of their areas so that small plans whose areas are small will be filtered out and highest polygons with big areas are considered as walls and they are represented in the 2D map by line segments that may present labeled discontinuities indicating the presence of door or window (see figure \ref{fig4}).\\

\begin{figure}[h!]
	\centering
	\includegraphics[width=10cm]{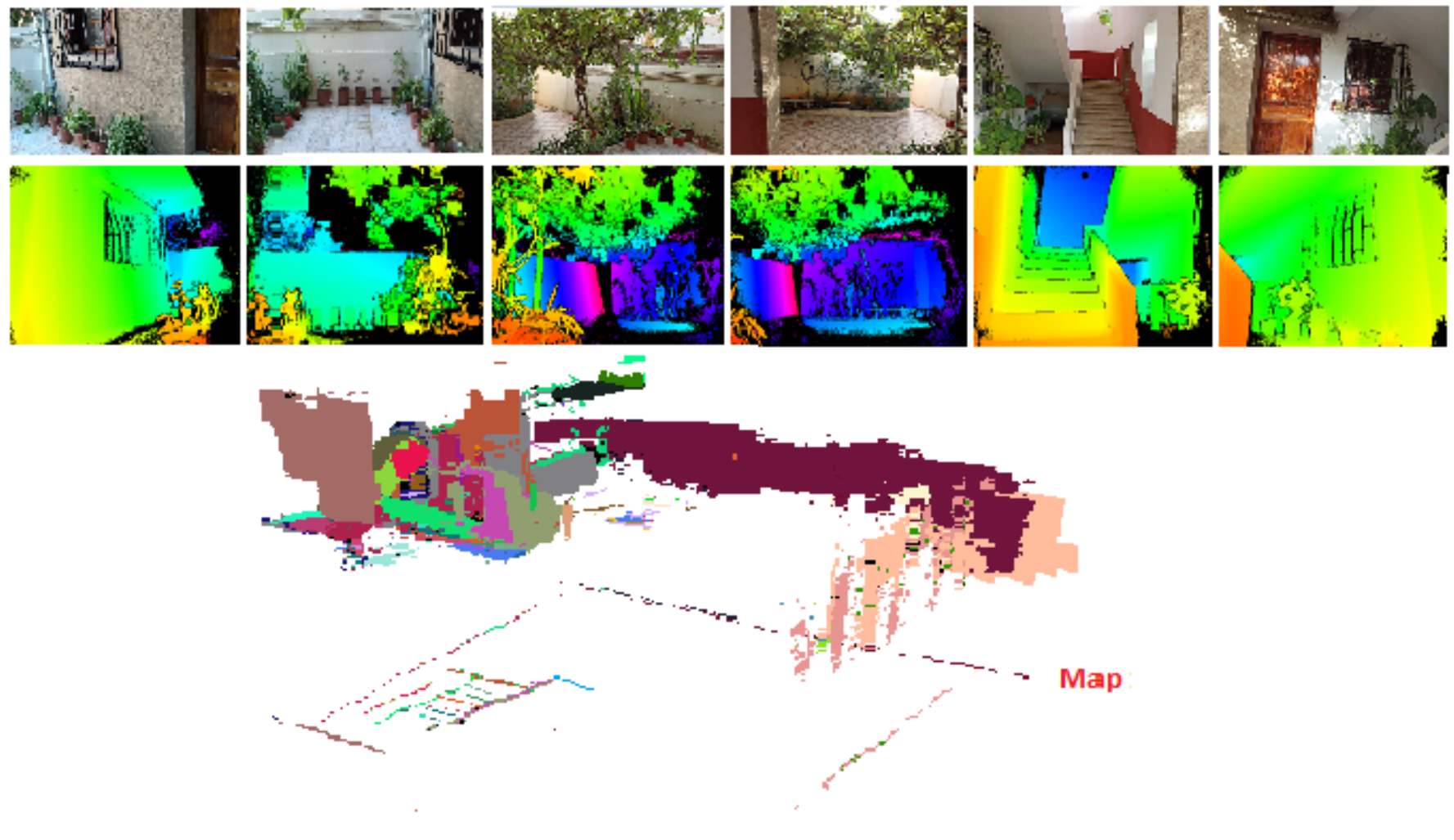}
	\caption{From top to bottom: RGB images, depth images, 3D model accompanied with 2D maps. The 2D map contains two apparent delimitations: The outside edges represent the scene borders, the rectangular shape illustrates the projection of the captured stair railing  whose parallel lines contained in its interior represent the staircase steps).}
     \label{fig5}
\end{figure}

- The stairs identified in the 3D map as set of parallel polygons regularly spaced (describing detected stair steps) parallel to  the ground are represented by a rectangular shape with parallel lines regularly spaced inside with a direction denoting if the detected stair steps in the 3D map are ascending or descending as illustrated in figure \ref{fig5} (for the descendant stair steps only treads polygons are detected in the 3D model however for ascending stairs, both treads and risers polygons are spotted).\\

\subsection{Descriptor computation}\label{subsect_scenedesc}

\begin{figure}[h!]
	\centering
	\includegraphics[width=8cm]{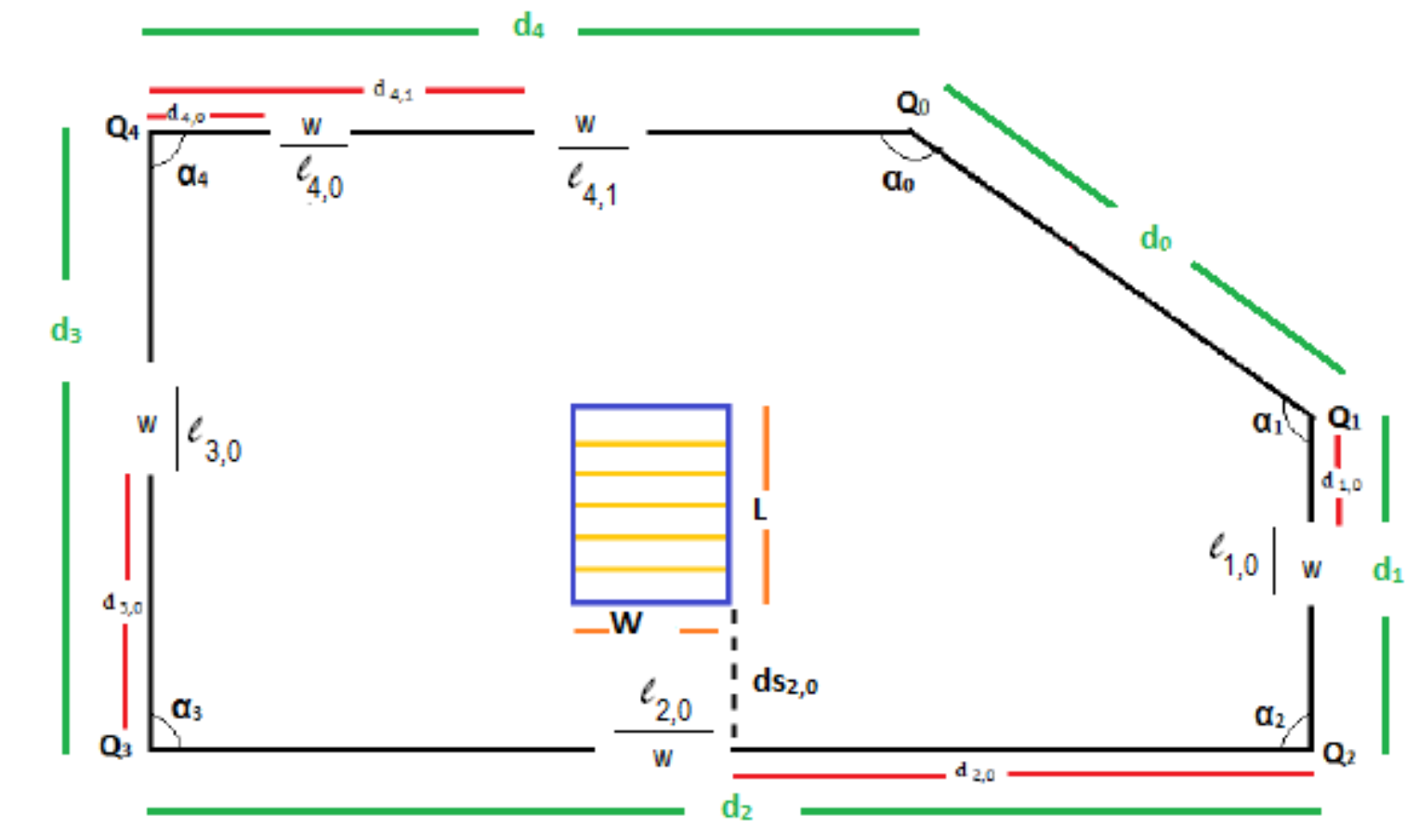}
	\caption{Example of 2D map computed in indoor environment.}
     \label{fig6}
\end{figure}

The extracted 2D map is described textually \cite{larabi2009} \cite{Aouat2010} and geometrically with a set of $n-uplets$. Each one, related to the vertex number $i$ contains the pair of (length of next segment and associated angle), a triplet of values (type of opening (door or window), its length and its distance from the vertex $i$) for each opening in segment connecting the vertices $i$ and $i+1$.
At the end of the descriptor, information about stairs are added. Each stair of the 2D map is coded with its type (ascendant or descendent), its dimensions(length, width), its distance from the nearest wall (see figure \ref{fig6}):

We note:\\
\begin{itemize}
    \item Corner $Q_k$.
    \item Segment line $L_{k}$ with length equal to $d_k$ demarcated between two successive corners $Q_k$ and $Q_{k+1}$, defining a wall in the 2D map. We specify that the segment length is set to zero when one of its two corners is not detected.
    \item Angles $\alpha_k$ between two successive line segments $L_{k-1}$ and $L_{k}$.
    \item $(t_{k,j}, O_{k,j}, d_{k,j})$: Indicates the presence of the $j^{th}$ opening on the wall $k$ of type $t_{k,j}$ (equal to 0 for windows and to $1$ for doors), of length $O_{k,j}$, at a distance $d_{k,j}$ from the corner $k$.
    \item $(t_{k,i}, l_{k,i}, w_{k,i}, ds_{k,i})$ indicates the presence of the $i^{th}$ stair nearest to wall $k$, of type $t_{k,i}=1$ (if the stairs are ascending otherwise $t_{k,i}=0$), of dimensions $(l_{k,i}, w_{k,i})$ at a distance $ds_{k,i}$ from the nearest wall $k$ (see figure \ref{fig6}).
\end{itemize}

If we describe the 2D map of figure \ref{fig6}, where the walls number $1, 3$ and $4$ contain respectively $1, 1$ and $2$ windows and the wall number $2$ contains a door, we obtain the following descriptor.\\

$D = \{Q_0(\alpha_0, d_0), Q_1(\alpha_1, d_1, (0, l_{1,0}, d_{1,0})),
Q_2(\alpha_2, d_2, (1, l_{2,0}, d_{2,0}),(1, L, W, ds_{2,0})) \\
Q_3(\alpha_3, d_3, (0, l_{3,0}, d_{3,0})), Q_4(\alpha_4, d_4, ( 0, l_{4,0}, d_{4,0}),\\
(0, l_{4,1}, d_{4,1})), \}$,

Algorithm \ref{algo1} summarizes the scene descriptor computation method.\\

\begin{algorithm}[t]
	\caption{Algorithm: The scene descriptor computation }
\label{algo1}
	\begin{algorithmic}[1]
		\renewcommand{\algorithmicrequire}{\textbf{Input:}}
		\renewcommand{\algorithmicensure}{\textbf{Output:}}
		\REQUIRE Video sequence of depth frames\\
		\ENSURE  $D_{s}$, the computed descriptor of the scene
		\FOR {Each frame}
            \STATE Extracting planar areas from selected frame;
            \STATE Clustering each set of planar areas that belong to the same plan into polygon;
            \STATE Merge the obtained polygons with polygons of the previous frames to build gradually the 3D map;
        \ENDFOR
           \STATE Determination of 2D map from computed 3D map;
          \STATE $D_{s}$ computation on the basis of determined 2D maps;

        \RETURN $D_{s}$
	\end{algorithmic}
\end{algorithm}

\section{The Place Recognition} \label{visual}

\begin{figure}
	\centering
	\includegraphics[width=7cm]{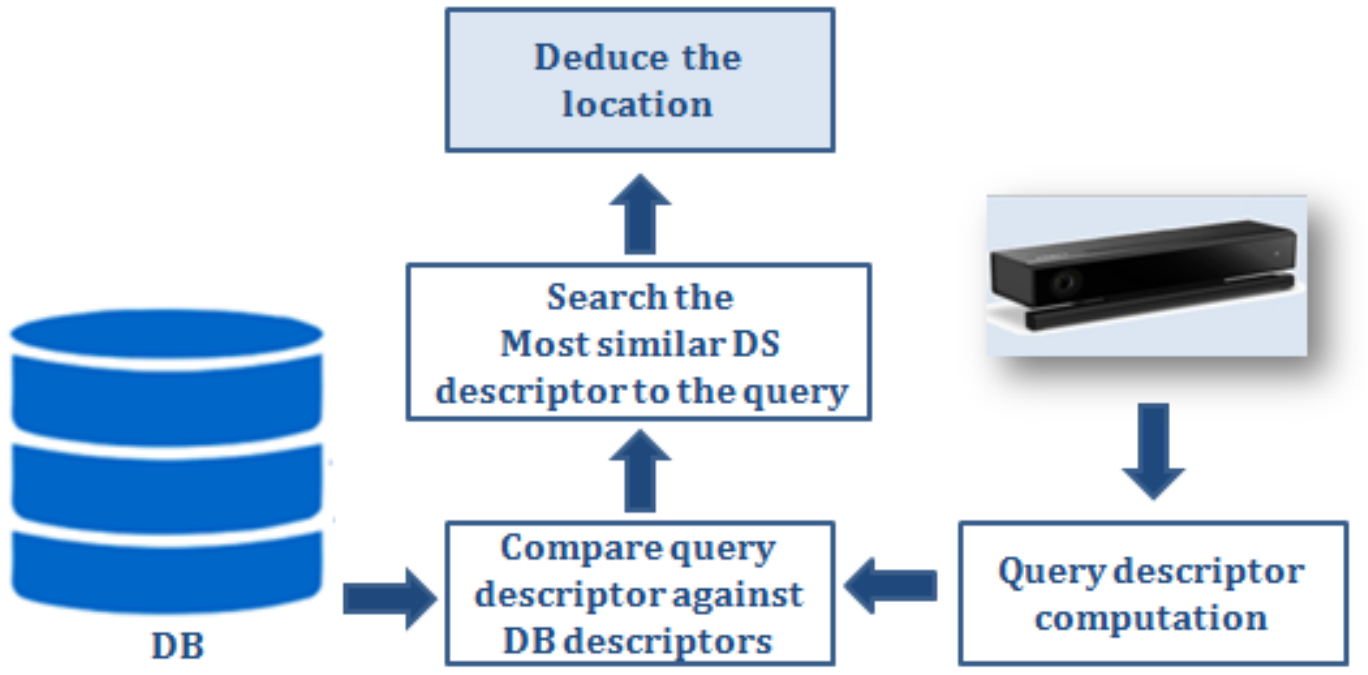}
	\caption{Diagram of the Place Recognition process that takes as input a scene depth video calculates the corresponding descriptor, compare it to the data-set descriptors and identify the location.}
     \label{fig7}
\end{figure}

The purpose of this section is the identification of the unknown location by retrieving the most similar database descriptor to the query  descriptor computed from query depth video of the captured scene using method of subsection \ref{subsect_scenedesc} (see figure \ref{fig7}).
We need in the first to built a places descriptors  database that will be used to retrieve the most similar descriptor.

\subsection{Dataset Construction} \label{dataset}

The public datasets may be classified into two classes. The first, encloses those that contain unsorted images list \cite{chen2011city} \cite{jegou2008hamming} suitable for evaluation of 2D based methods and neural network based methods, they furnish query images captured under various conditions. The second for sorted images list \cite{sturm12iros} \cite{kendall2015posenet} destined for 3D based methods assessment. They do  not consider big changes between query and database images due to relying on feature matching for ground truth generation. As none of these data-sets can be used to evaluate our proposed system based on coarse scene information because the evaluation of this later requires a set of videos (sorted images list) taken under various acquisition conditions and these datasets do not furnish the required data, we built our dataset \cite{ibelaidenbenchmark}.

\begin{figure}
	\centering
	\includegraphics[width=10cm]{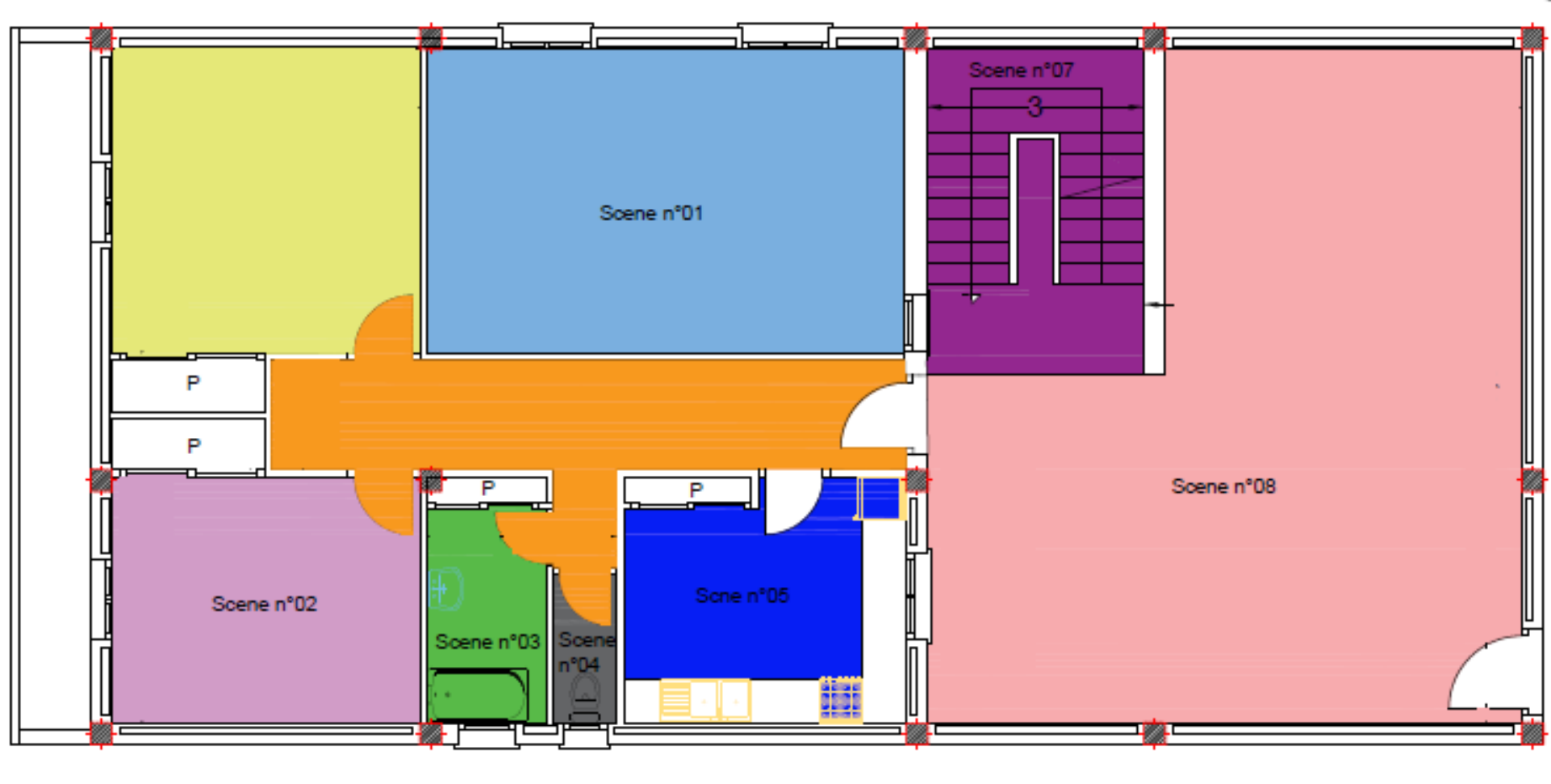}
	\caption{ Experimental environment floor-plan.}
     \label{fig8}
\end{figure}

This is made by capturing depth videos of different indoor scenes using the depth camera.
Figure \ref{fig8} illustrates an example of indoor working environment having scenes of various geometric forms (rectangular (scene 01), squarish (scene 00), T (scene 06), L (scene 08) and N (scene 07) shapes) and dimensions (scenes of small surfaces $(10m^2)$, medium $(30m^2)$ and large$(70m^2)$).

\begin{figure}
	\centering
	\includegraphics[width=8cm]{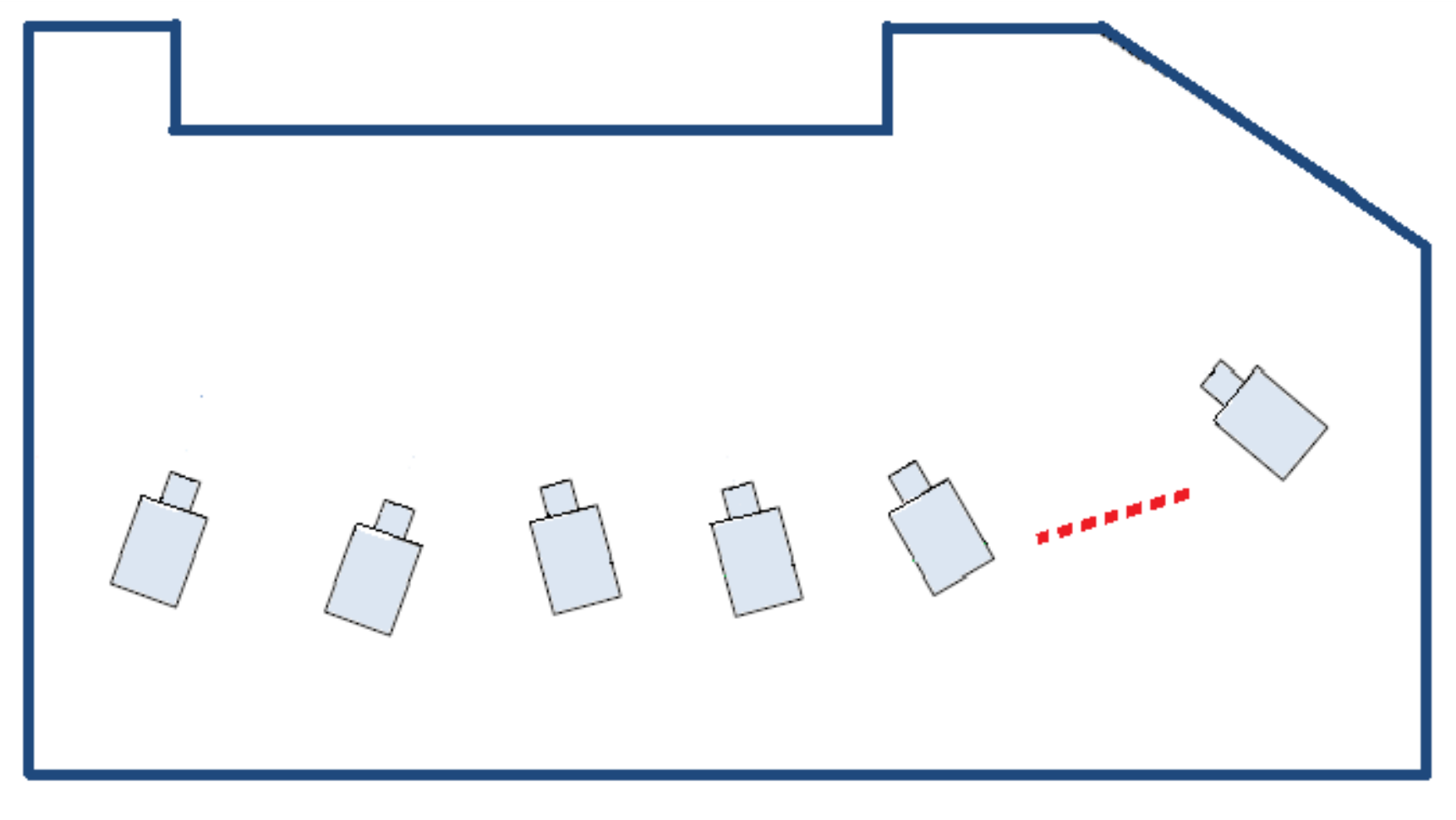}
	\caption{Data acquisition in scenes of intricate structure and big dimensions.}
    \label{fig9}
\end{figure}

A depth camera tied up to a cart was displaced in each scene, so that it is fully overcasted in its corresponding video. In scenes of simple structure and  small dimensions the camera was merely surrounded; whereas in scenes of intricate structures and big dimensions it was rotated and translated in order to cover all scene details see figure \ref{fig9}.


Queries depth videos covering different scene parts from different positions and under various conditions were likewise captured  in order to be used for evaluation.
Each recorded video is used to compute the scene descriptor $D_{scene}$ (using the explained method in subsection \ref{subsect_scenedesc}) then the couple $\{D_{scene}, Location\}$ ($Location$ indicate the identification of scene) is stored in the database accompanied with RGB images (saved for evaluation to compare the results of the state of the art method against those of the proposed system).

\subsection{Descriptors Matching} \label{match}

Let $D_{q}= \{Q_{k}(\alpha_{k}, d_{k}, (t_{k,j}, l_{k,j}, d_{k,j}), (t_{k,i}, L_{k,i}, W_{k,i}, ds_{k,i})\}$
be the descriptor of unknown scene (see subsection \ref{subsect_scenedesc}).

Let $D_p$ be the descriptor of a scene from the dataset.\\
$D_{p}= \{Q'_{k}(\alpha'_{k}, d'_{k}, (t'_{k,j}, l'_{k,j}, d'_{k,j})),(t'_{k,i}, L'_{k,i}, W'_{k,i}, ds'_{k,i})\}$\\

\begin{figure}
	\centering
	\includegraphics[width=8cm]{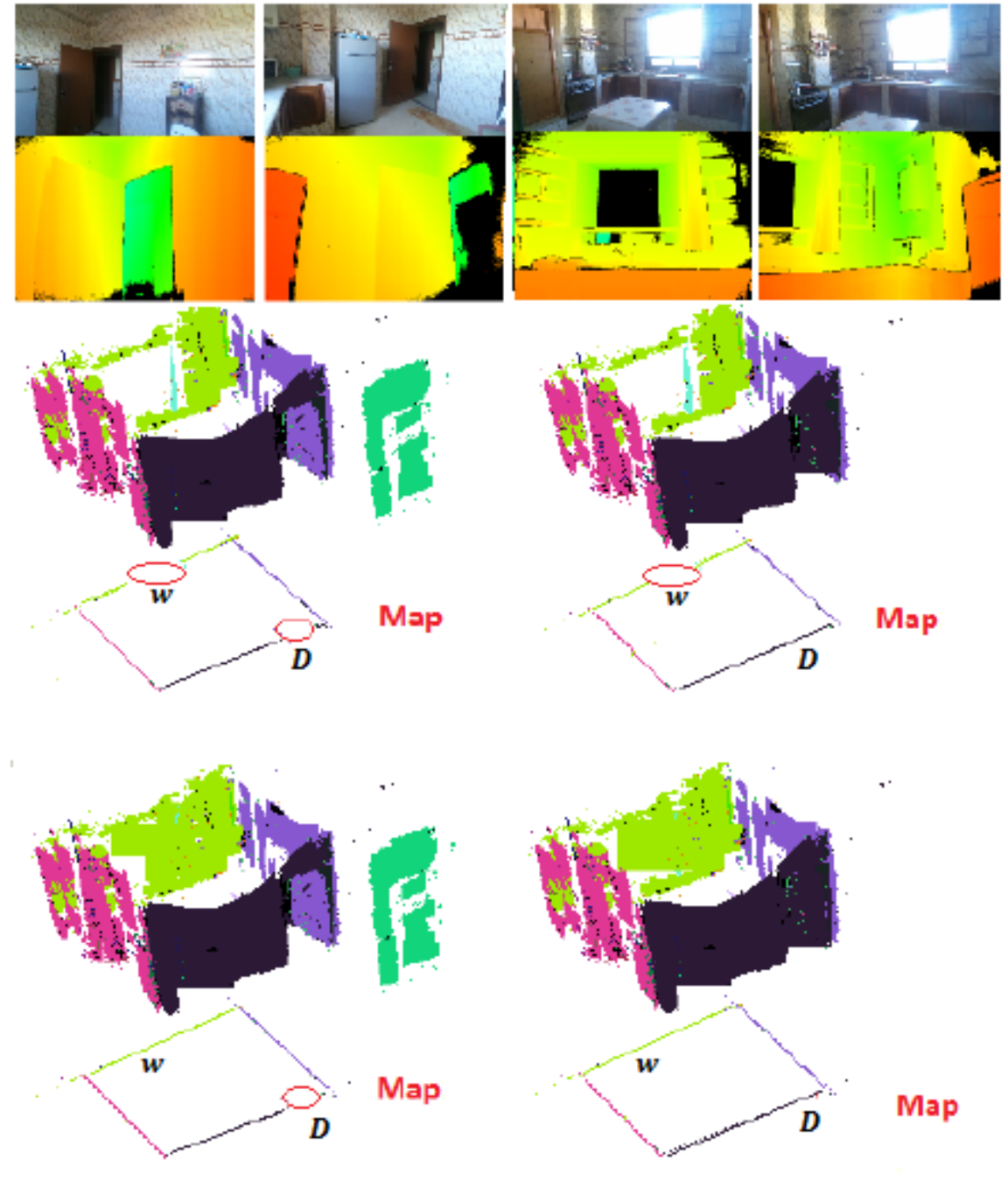}
	\caption{ The 2D maps constructed for the same indoor scene considering the different possible states of its apertures: the door and window (open, close).}
     \label{fig10}
\end{figure}

As the starting point of dataset descriptor for the comparison is unknown, the difference measure between $D_{q}$ and $D_p$ is computed for all possible starting points $j$ considering the clockwise and anticlockwise directions. For each dataset descriptor in sequential (whose corners number is greater or equal to that of the query descriptor) a difference measure is computed and the minimal measure is selected and saved for the considered descriptor. The one having a minimum difference measure is considered as appropriate to the query and its location is transferred to that of the query. Note that all  possible states of apertures (open, close) were considered for comparison as illustrated by figure \ref{fig10}(the openings are circled in red). These steps are summarized in algorithm \ref{algo2}.

\begin{algorithm}[t]
	\caption{Place Recognition Algorithm}
\label{algo2}
	\begin{algorithmic}[1]
		\renewcommand{\algorithmicrequire}{\textbf{Input:}}
		\renewcommand{\algorithmicensure}{\textbf{Output:}}
		\REQUIRE $D_{q}$: the query descriptor having $m$ corners\\
		\ENSURE  $Location$ the query descriptor location\\
		\FOR {each $\; D_{s} \in \mathbb{DS}$}
           \FOR {each $\; Starting \; corner\; Q_{j}^M \; of \; D_{s}$ considering the clock wise and anti clock wise directions}
              \STATE $Compute Similarity Measure \; Sim(Q_{0}^{R},  Q_{j}^{M})$ ;
              \STATE $Select \; D^* \; such \; that \; D(Q_{0}^{R},  Q_{j}^{M}) \; is \; minimal$ ;
            \ENDFOR
            \STATE $Save\; the \; location \; of  \; D_{s}  \;  \; if \; it \;  is \;  more \;  similar  \; to  \; D_{q}$ ;
         \ENDFOR

        \RETURN $Location \; of \; the \; query  \;descriptor$
	\end{algorithmic}
\end{algorithm}



\begin{multline*}
\label{eqs}
Sim(D_q, D_p)=
\sum_{k=0}^{nq} (\beta_1 * \| \alpha_k - \alpha'_{k}\| +  \beta_2 * \|  d_{k} - d'_{k}\| +
\sum_{p=0}^{no_k} (\beta_3 * \|t_{k,p} - t'_{k,p}\| + \\ \beta_2 *( \|d_{k,p} - d'_{k,p}\| +  \|l_{k,p} - l'_{k,p}\|))+
\sum_{i=0}^{ns_k} (\beta_2 * (\| W_{k,i} - W'_{k,i}\| + \| L_{k,i} - L'_{k,i}\| +\\\| ds_{k,i} - ds'_{k,i}\|) + \beta_3 * \| t_{k,i} - t'_{k,i}\|))
\end{multline*}

$\beta_1$, $\beta_2$ and $\beta_3$ are the coefficients  associated to each feature according to its importance in the determination of  similarity measure between places (they were chosen on the basis of experimentation).

\section{Experiments}\label{exp}

Through this section we assess the performance and effectiveness of the proposed system which was implemented using c++ programming language and executed in a laptop with an Intel Core i5-5200U  CPU @ 2.20 GHz and 4 GB RAM.
The first subsection, presents place recognition accuracy  verification, the second assess the place recognition approach and  compares it to the relevant state of the art methods Fabmap \cite{cummins2011appearance} and FastABLE \cite{nowicki2017real}. Finally, the third  exhibits the computational complexity results.

\subsection{Exactness of scenes descriptors}

\begin{figure}
	\centering
	\includegraphics[width=12cm]{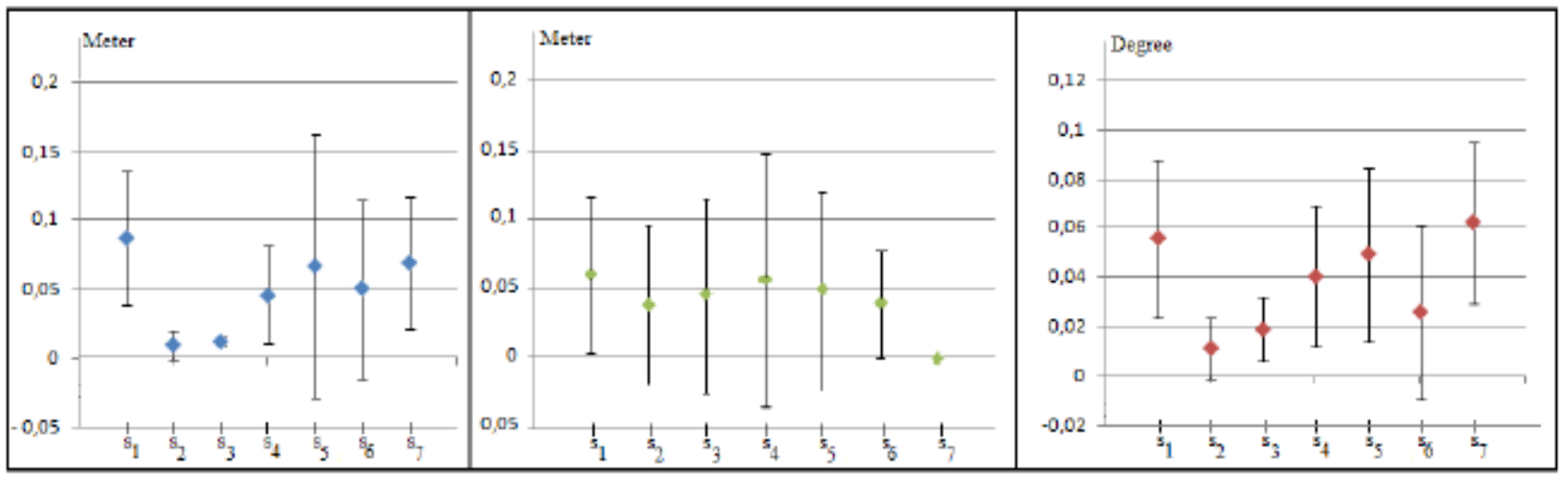}
	\caption{From left to right: average errors and standard deviations for calculated wall lengths, apertures and angles for seven scenes $(s_1, ..., s_7)$.}
     \label{g1}
\end{figure}

As the proposed descriptors represent plane views of the scenes, we have evaluated their precision by comparing the derived 2D maps from which these descriptors were computed against real plane views.
Figure \ref{g1} evinces the average error and standard deviation of calculated dimensions, angles and overtures in 2D maps of seven scenes sample.

\begin{figure}
	\centering
	\includegraphics[width=12cm]{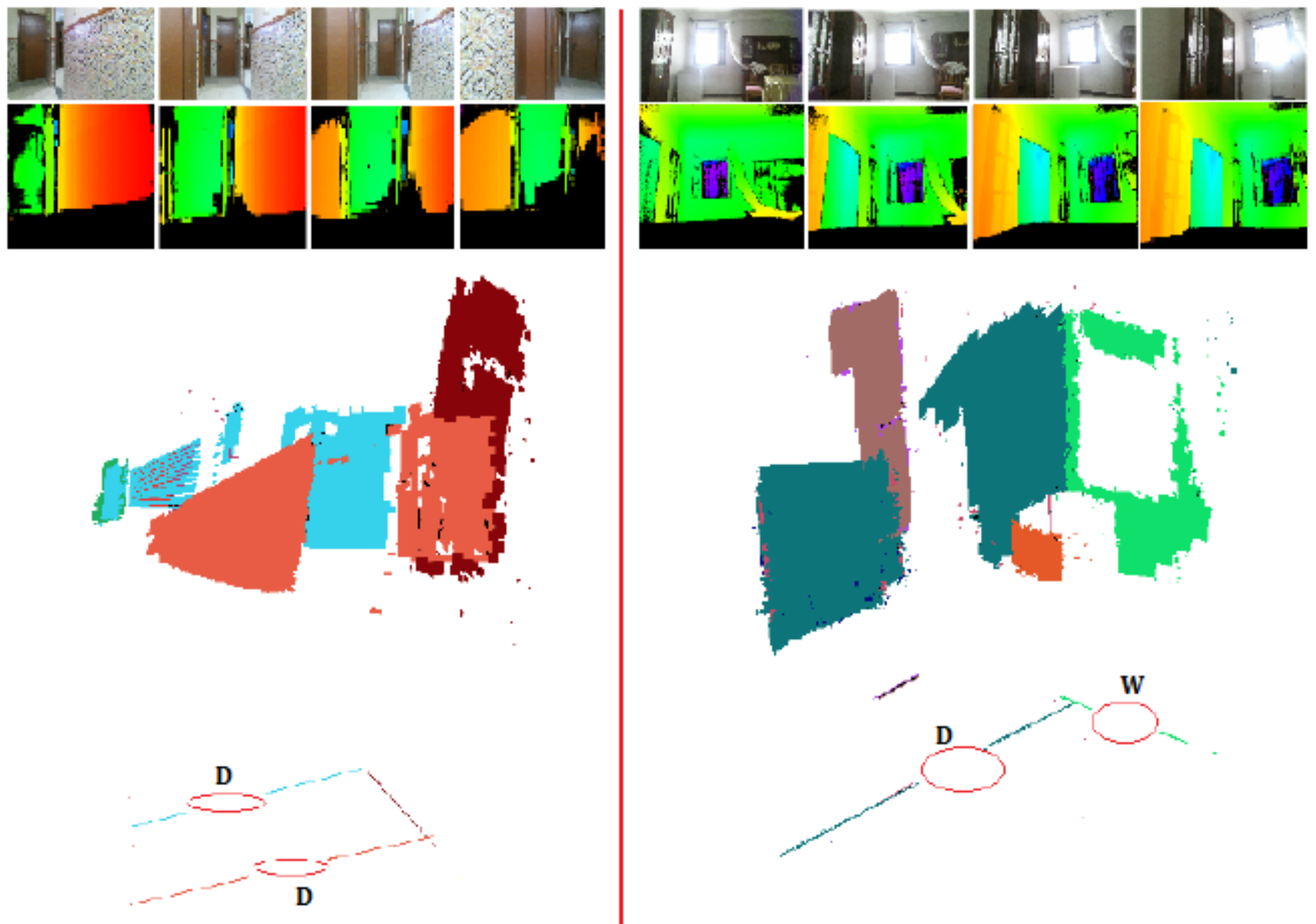}
	\caption{Overtures of doors (left) and windows (right) have been
correctly identified and distinguished.}
     \label{fig12}
\end{figure}

Compared to the ground truth, computed measures from 2D maps are accurate (figure \ref{g1}), moreover the doors overtures have been correctly identified and distinguished from those that represent windows as shown in figure \ref{fig12} where the orange plan in the left side of this later was not represented in the 2D map because only walls planes were projected.
Note also that apertures dimensions are independent of the sash opening (in whole or in partial) because they are defined as overtures of the same plane.

\subsection{Evaluation of Place Recognition Approach}

The dataset presented in \cite{ibelaidenbenchmark} was used to evaluate the system performance. It includes two sets of depth videos. The first, contains those that was used to construct model descriptors while the second for test data which may be subdivided into eight subsets considering light variations, scenery changes, appearance and disappearance of movable objects and persons (like summarised in table \ref{tb}) in order to investigate the effect of  changing acquisition conditions.

\begin{table}[ht]
    \centering
    \begin{tabular}{|c|c|c|c|}
    \hline
          Scene & lighting variation & Dynamic objects & Scenery changes\\
    \hline
    Subset 1 & No & No & No\\
    \hline
     Subset 2 & No & No & Yes \\
    \hline
    Subset 3 & No & Yes & No\\
    \hline
    Subset 4 & No & Yes & Yes\\
    \hline
    Subset 5 & Yes & No & No\\
    \hline
    Subset 6 & Yes & No & Yes\\
    \hline
    Subset 7 & Yes & Yes & No\\
    \hline
    Subset 8 & Yes & Yes & Yes\\
    \hline
        \end{tabular}
    \caption{Subsets of data test.}
    \label{tb}
\end{table}

\begin{figure}
	\centering
	\includegraphics[width=11cm]{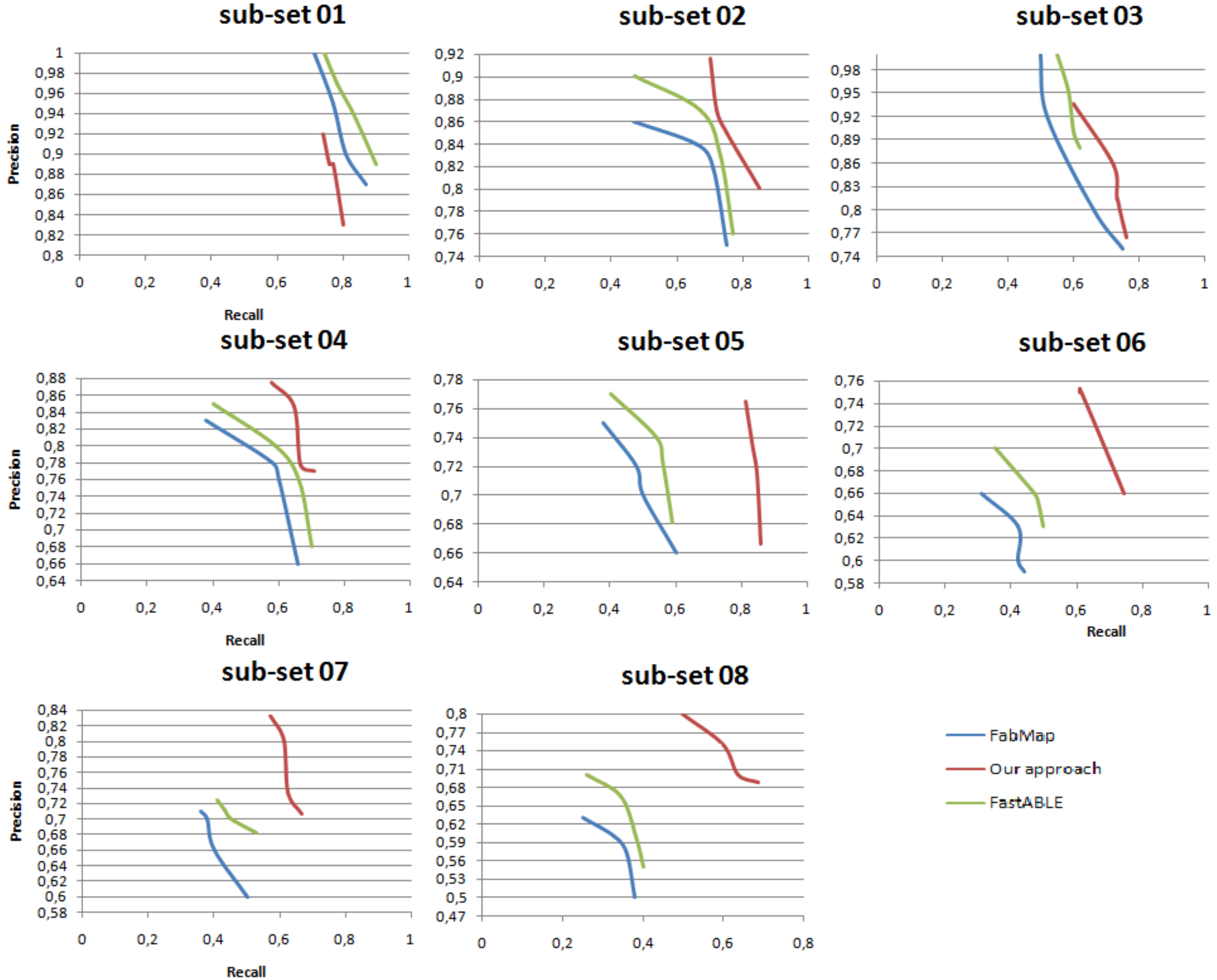}
	\caption{Precision-recall curves of the proposed method, FastABLE and FABMAP for each subset of test data.}
     \label{figure13}
\end{figure}

The propounded method that takes as input a sequence of depth frames was compared to FastABLE algorithm \cite{nowicki2017real} (uses a sequence of RGB images) and Fast Appearance Based Mapping (FABMAP) approach \cite{cummins2011appearance} (utilizes Rgb image). Both FastABLE and FABMAP algorithms available in open source \cite{glover2012openfabmap} were evaluated on the constructed dataset in order to establish a reliable comparison. The precision-recall curves of the three approaches are presented in figure \ref{figure13} for each subset of test data and in table \ref{tb3} the F1 scores are given.

\textcolor{red}{
\begin{table}[]
    \centering
    \begin{tabular}{|c|c|c|c|}
    \hline  Subset & Proposed approach F1 & FastABLE F1 & Fabmap F1\\
    \hline  Subset 1 & \textbf{0.8210} &          0.8762 & 0.8549 \\
    \hline  Subset 2 & \textbf{0.8071} &         0.7399  & 0.7220\\
    \hline  Subset 3 & \textbf{0.7669} &          0.7213 & 0.7146 \\
    \hline  Subset 4 & \textbf{0.7243} &          0.6681 & 0.6406 \\
    \hline  Subset 5 & \textbf{0.7739} &          0.6083 & 0.5789 \\
    \hline  Subset 6 & \textbf{0.6890} &          0.5353 & 0.4788 \\
    \hline  Subset 7 &\textbf{ 0.6852} &          0.5528 & 0.5080 \\
    \hline  Subset 8 & \textbf{0.6639} &          0.4466 & 0.4193 \\
    \hline
    \end{tabular}
    \caption{ F1 scores of the proposed system, FastABLE and Fabmap.}
    \label{tb3}
\end{table}
}

\begin{figure}
	\centering
	\includegraphics[width=7cm]{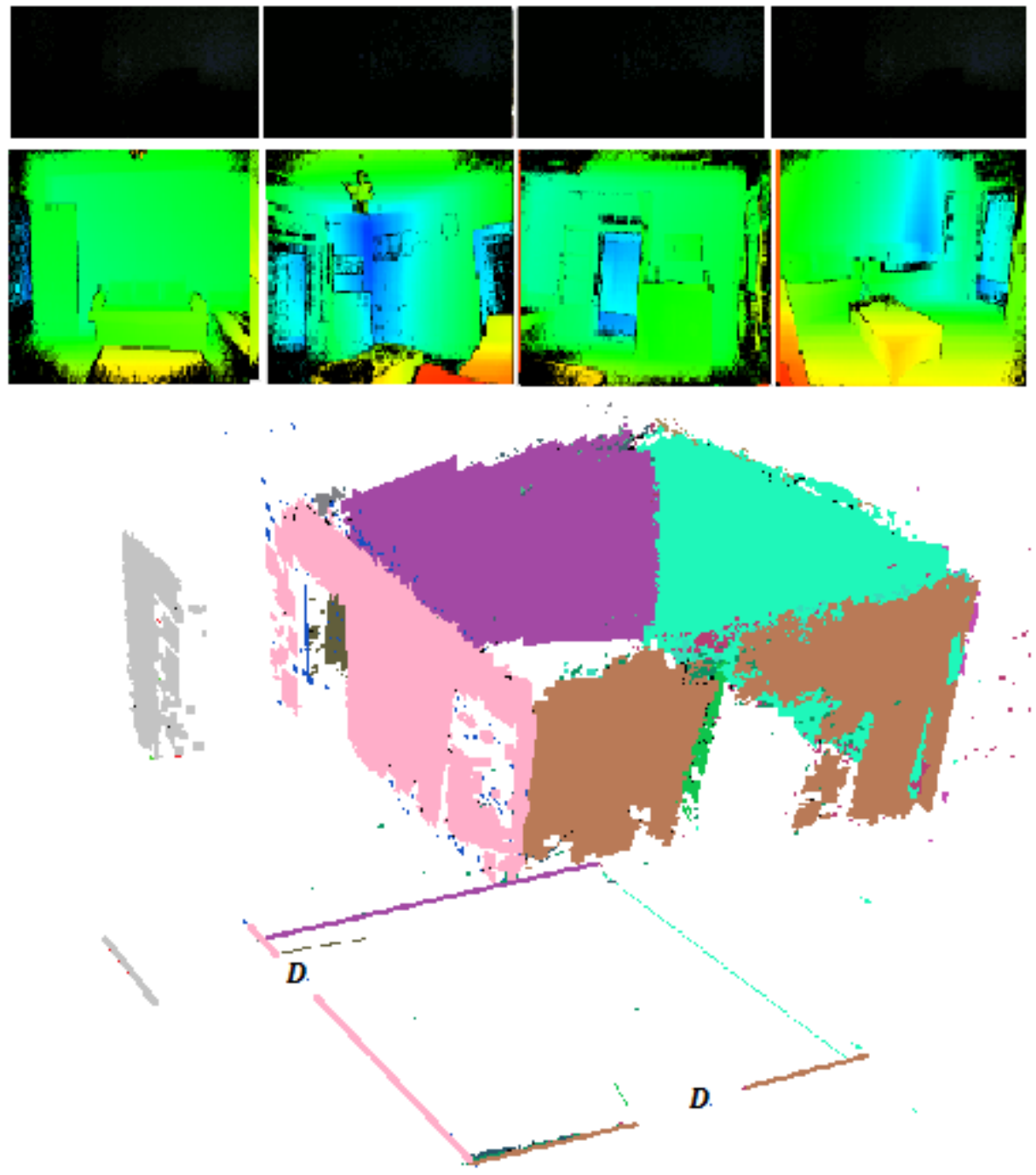}
	\caption{3D and 2D maps computed in very low light scene.}
     \label{low}
\end{figure}

The exposed results in figure \ref{figure13} and table \ref{tb3} show that:\\
- For the first subset, FastABLE and FABMAP  offer better results than the proposed approach; \\
- In case of scenery changes, the place recognition results decrease in both state of the art methods and still stable in our approach;\\
- Concerning sensitivity to light changes, FastABLE and FABMAP are more responsive than the proposed method, figure \ref{low} shows 3D and 2D maps constructed in very low light scene in order to affirm that our system remains accurate even in very low light.\\
- The dynamic objects and persons in locations decrease slightly the place recognition results in the three methods;\\
- When scenes undergo big changes in appearance due to combination of aforementioned acquisition conditions (subset 4, subset 6, subset 7 and subset 8), the accuracy of the proposed approach is higher than that of FastABLE and FABMAP.

\subsection{The proposed system computational complexity}

Our system consists of the following processing stages:
\begin{itemize}
    \item Planar regions extraction from depth frames;
    \item Merging of planar regions belonging to the same plan  into polygon;
    \item 3D map Construction;
    \item 2D map derivation;
    \item Scene description;
    \item place recognition.
\end{itemize}

The computational complexity of each step is indicated in table \ref{tb2}, the measurements were taken from the execution of the proposed algorithm on different scenes (with a total of 1000 images per scene). Note that the multi-threading and GPUs were not used in the system implementation.

\begin{table}[]
    \centering
    \begin{tabular}{|c|c|c|c|}
    \hline Stage & Max time & Average time& Min time\\
\hline Planar regions extraction & 0.0107 & 0.0067 & 0.0028  \\
\hline Polygon construction  & 0.0042  & 0.0021 & 0.0001 \\
\hline 3D map Construction  & 0.0309 &  0.0015&  0.0004 \\
\hline 2D map derivation  & 0.0080 & 0.0065 & 0.0050 \\
\hline Scene description  & 1.1222 & 1.3667 & 1.6112 \\
\hline Place recognition  &  0.0005 & 0.0004 & 0.0003 \\
\hline Total & 1.1658 & 1.3772 & 1.6170\\
\hline
  \end{tabular}
    \caption{Computational complexity of the proposed algorithm.}
    \label{tb2}
\end{table}

Despite slight high of the system place recognition time in comparison to FastABLE and Fabmap, The proposed approach in this paper is more suitable for applications where robustness to real scenarios (illumination variation, scenery modifications) is required.

\section{Conclusion and future works}\label{conclusion}

We have propounded a new place recognition method whose dataset does not require updates because it contains architecture based scene descriptors reflecting the non rearrangeable scene parts.
The conducted experiments show the descriptors accuracy and the benefit of their use on place recognition especially in case of challenging acquisition conditions where the propounded system outperforms the state of the art methods FastABLE \cite{nowicki2017real} and Fabmap \cite{cummins2011appearance}. However a confusion is expected in the proposed approach when different scenes present exactly the same architecture that's why we plan to study the possibility of this inconvenient elimination by the include of not movable  objects in scene descriptor and analyze their impact on place recognition results.

\bibliographystyle{elsarticle-num}

\begin{thebibliography}{0}

\bibitem{liang2013image}
J.~Z. Liang, N.~Corso, E.~Turner, A.~Zakhor, Image based localization in indoor
  environments, in: 2013 Fourth International Conference on Computing for
  Geospatial Research and Application, IEEE, 2013, pp. 70--75.

\bibitem{acharya2019bim}
D.~Acharya, M.~Ramezani, K.~Khoshelham, S.~Winter, Bim-tracker: A model-based
  visual tracking approach for indoor localisation using a 3d building model,
  ISPRS Journal of Photogrammetry and Remote Sensing 150 (2019) 157--171.

\bibitem{lowry2015visual}
S.~Lowry, N.~S{\"u}nderhauf, P.~Newman, J.~J. Leonard, D.~Cox, P.~Corke, M.~J.
  Milford, Visual place recognition: A survey, IEEE Transactions on Robotics
  32~(1) (2015) 1--19.

\bibitem{Wohlkinger2011EnsembleOS}
W.~Wohlkinger, M.~Vincze, Ensemble of shape functions for 3d object
  classification, 2011 IEEE International Conference on Robotics and
  Biomimetics (2011) 2987--2992.

\bibitem{rusu2010fast}
R.~B. Rusu, G.~Bradski, R.~Thibaux, J.~Hsu, Fast 3d recognition and pose using
  the viewpoint feature histogram, in: 2010 IEEE/RSJ International Conference
  on Intelligent Robots and Systems, IEEE, 2010, pp. 2155--2162.

\bibitem{aldoma2011cad}
A.~Aldoma, M.~Vincze, N.~Blodow, D.~Gossow, S.~Gedikli, R.~B. Rusu, G.~Bradski,
  Cad-model recognition and 6dof pose estimation using 3d cues, in: 2011 IEEE
  international conference on computer vision workshops (ICCV workshops), IEEE,
  2011, pp. 585--592.

\bibitem{chen2018indoor}
Y.~Chen, R.~Chen, M.~Liu, A.~Xiao, D.~Wu, S.~Zhao, Indoor visual positioning
  aided by cnn-based image retrieval: Training-free, 3d modeling-free, Sensors
  18~(8) (2018) 2692.

\bibitem{sizikova2016enhancing}
E.~Sizikova, V.~K. Singh, B.~Georgescu, M.~Halber, K.~Ma, T.~Chen, Enhancing
  place recognition using joint intensity-depth analysis and synthetic data,
  in: European Conference on Computer Vision, Springer, 2016, pp. 901--908.

\bibitem{song2019learning}
X.~Song, S.~Jiang, L.~Herranz, C.~Chen, Learning effective rgb-d
  representations for scene recognition, IEEE Transactions on Image Processing
  28~(2) (2019) 980--993.

\bibitem{koskela2014convolutional}
M.~Koskela, J.~Laaksonen, Convolutional network features for scene recognition,
  in: Proceedings of the 22nd ACM international conference on Multimedia, ACM,
  2014, pp. 1169--1172.

\bibitem{taira2018inloc}
H.~Taira, M.~Okutomi, T.~Sattler, M.~Cimpoi, M.~Pollefeys, J.~Sivic, T.~Pajdla,
  A.~Torii, Inloc: Indoor visual localization with dense matching and view
  synthesis, in: Proceedings of the IEEE Conference on Computer Vision and
  Pattern Recognition, 2018, pp. 7199--7209.

\bibitem{oliva2001modeling}
A.~Oliva, A.~Torralba, Modeling the shape of the scene: A holistic
  representation of the spatial envelope, International journal of computer
  vision 42~(3) (2001) 145--175.

\bibitem{cupec2015place}
R.~Cupec, E.~K. Nyarko, D.~Filko, A.~Kitanov, I.~Petrovi{\'c}, Place
  recognition based on matching of planar surfaces and line segments, The
  International Journal of Robotics Research 34~(4-5) (2015) 674--704.

\bibitem{Zou2017Complete3S}
C.~Zou, Z.~Li, D.~Hoiem, Complete 3d scene parsing from single rgbd image, CoRR
  abs/1710.09490.

\bibitem{ma2018indoor}
W.~Ma, H.~Xiong, X.~Dai, X.~Zheng, Y.~Zhou, An indoor scene recognition-based
  3d registration mechanism for real-time ar-gis visualization in mobile
  applications, ISPRS International Journal of Geo-Information 7~(3) (2018)
  112.

\bibitem{lowe1999object}
D.~G. Lowe, et~al., Object recognition from local scale-invariant features.,
  in: iccv, Vol.~99, 1999, pp. 1150--1157.

\bibitem{rublee2011orb}
E.~Rublee, V.~Rabaud, K.~Konolige, G.~R. Bradski, Orb: An efficient alternative
  to sift or surf., in: ICCV, Vol.~11, Citeseer, 2011, p.~2.

\bibitem{bay2006surf}
H.~Bay, T.~Tuytelaars, L.~Van~Gool, Surf: Speeded up robust features, in:
  European conference on computer vision, Springer, 2006, pp. 404--417.

\bibitem{1245}
R.~B. Rusu, Z.~C. Marton, N.~Blodow, M.~Beetz, Learning informative point
  classes for the acquisition of object model maps, in: Proceedings of the 10th
  International Conference on Control, Automation, Robotics and Vision
  (ICARCV), Hanoi, Vietnam, 2008.

\bibitem{ye2017place}
Y.~Ye, T.~Cieslewski, A.~Loquercio, D.~Scaramuzza, Place recognition in
  semi-dense maps: Geometric and learning-based approaches, in: Proc. Brit.
  Mach. Vis. Conf., 2017, pp. 72--1.

\bibitem{deretey2015visual}
E.~Deretey, M.~T. Ahmed, J.~A. Marshall, M.~Greenspan, Visual indoor
  positioning with a single camera using pnp, in: 2015 International Conference
  on Indoor Positioning and Indoor Navigation (IPIN), IEEE, 2015, pp. 1--9.

\bibitem{wen2019efficient}
H.~Wen, R.~Clark, S.~Wang, X.~Lu, B.~Du, W.~Hu, N.~Trigoni, Efficient indoor
  positioning with visual experiences via lifelong learning, IEEE Transactions
  on Mobile Computing 18~(4) (2019) 814--829.

\bibitem{code3}
M.~Uddin, Scene classification using localized histogram of oriented gradients
  method, International Journal of Computer (IJC) 20~(1) (2016) 13--18.

\bibitem{sahdev2016indoor}
R.~Sahdev, J.~K. Tsotsos, Indoor place recognition system for localization of
  mobile robots, in: 2016 13th Conference on Computer and Robot Vision (CRV),
  IEEE, 2016, pp. 53--60.

\bibitem{Moon2016CaPSuLeAC}
Y.~Moon, S.~Noh, D.~Park, C.~Luo, A.~Shrivastava, S.~Hong, K.~V. Palem,
  Capsule: A camera-based positioning system using learning, 2016 29th IEEE
  International System-on-Chip Conference (SOCC) (2016) 235--240.

\bibitem{qiao2017visual}
Y.~Qiao, Z.~Zhang, Visual localization by place recognition based on
  multifeature (d-$\lambda$lbp, Journal of Sensors 2017.

\bibitem{zheng2017visual}
Y.~Zheng, P.~Luo, S.~Chen, J.~Hao, H.~Cheng, Visual search based indoor
  localization in low light via rgb-d camera, World Academy of Science,
  Engineering and Technology, International Journal of Computer, Electrical,
  Automation, Control and Information Engineering 11~(3) (2017) 349--352.

\bibitem{knopp2010avoiding}
J.~Knopp, J.~Sivic, T.~Pajdla, Avoiding confusing features in place
  recognition, in: European Conference on Computer Vision, Springer, 2010, pp.
  748--761.

\bibitem{torii2013visual}
A.~Torii, J.~Sivic, T.~Pajdla, M.~Okutomi, Visual place recognition with
  repetitive structures, in: Proceedings of the IEEE conference on computer
  vision and pattern recognition, 2013, pp. 883--890.

\bibitem{feng2017visual}
G.~Feng, L.~Ma, X.~Tan, Visual map construction using rgb-d sensors for
  image-based localization in indoor environments, Journal of Sensors 2017.

\bibitem{chen2017vision}
K.-W. Chen, C.-H. Wang, X.~Wei, Q.~Liang, C.-S. Chen, M.-H. Yang, Y.-P. Hung,
  Vision-based positioning for internet-of-vehicles, IEEE Transactions on
  Intelligent Transportation Systems 18~(2) (2017) 364--376.

\bibitem{gao2019mobile}
M.~Gao, M.~Yu, H.~Guo, Y.~Xu, Mobile robot indoor positioning based on a
  combination of visual and inertial sensors, Sensors 19~(8) (2019) 1773.

\bibitem{lepetit2009epnp}
V.~Lepetit, F.~Moreno-Noguer, P.~Fua, Epnp: An accurate o (n) solution to the
  pnp problem, International journal of computer vision 81~(2) (2009) 155.

\bibitem{sattler2017efficient}
T.~Sattler, B.~Leibe, L.~Kobbelt, Efficient \& effective prioritized matching
  for large-scale image-based localization, IEEE transactions on pattern
  analysis and machine intelligence 39~(9) (2017) 1744--1756.

\bibitem{li2010location}
Y.~Li, N.~Snavely, D.~P. Huttenlocher, Location recognition using prioritized
  feature matching, in: European conference on computer vision, Springer, 2010,
  pp. 791--804.

\bibitem{he2015spatial}
K.~He, X.~Zhang, S.~Ren, J.~Sun, Spatial pyramid pooling in deep convolutional
  networks for visual recognition, IEEE transactions on pattern analysis and
  machine intelligence 37~(9) (2015) 1904--1916.

\bibitem{oertel2020augmenting}
A.~Oertel, T.~Cieslewski, D.~Scaramuzza, Augmenting visual place recognition
  with structural cues, arXiv preprint arXiv:2003.00278.

\bibitem{fan2020visual}
H.~Fan, Y.~Zhou, A.~Li, S.~Gao, J.~Li, Y.~Guo, Visual localization using
  semantic segmentation and depth prediction, arXiv preprint arXiv:2005.11922.

\bibitem{irschara2009structure}
A.~Irschara, C.~Zach, J.-M. Frahm, H.~Bischof, From structure-from-motion point
  clouds to fast location recognition, in: 2009 IEEE Conference on Computer
  Vision and Pattern Recognition, IEEE, 2009, pp. 2599--2606.

\bibitem{kendall2015posenet}
A.~Kendall, M.~Grimes, R.~Cipolla, Posenet: A convolutional network for
  real-time 6-dof camera relocalization, in: Proceedings of the IEEE
  international conference on computer vision, 2015, pp. 2938--2946.

\bibitem{deng2017unsupervised}
Z.~Deng, S.~Todorovic, L.~J. Latecki, Unsupervised object region proposals for
  rgb-d indoor scenes, Computer Vision and Image Understanding 154 (2017)
  127--136.

\bibitem{finman2015toward}
R.~Finman, L.~Paull, J.~J. Leonard, Toward object-based place recognition in
  dense rgb-d maps, in: ICRA Workshop Visual Place Recognition in Changing
  Environments, Seattle, WA, Vol.~76, 2015.

\bibitem{boniardi2017robust}
F.~Boniardi, T.~Caselitz, R.~K{\"u}mmerle, W.~Burgard, Robust lidar-based
  localization in architectural floor plans, in: 2017 IEEE/RSJ International
  Conference on Intelligent Robots and Systems (IROS), IEEE, 2017, pp.
  3318--3324.

\bibitem{boniardi2019robot}
F.~Boniardi, A.~Valada, R.~Mohan, T.~Caselitz, W.~Burgard, Robot localization
  in floor plans using a room layout edge extraction network, arXiv preprint
  arXiv:1903.01804.

\bibitem{ibelaiden2020scene}
F.~Ibelaiden, B.~Sayah, S.~Larabi, Scene description
from depth images for visually positioning, in: 020 1st International Conference on Communications,
  Control Systems and Signal Processing (CCSSP), IEEE, 2020, pp. 101--106.

\bibitem{ibelaidenbenchmark}
F.~Ibelaiden, S.~Larabi, A benchmark for visual positioning from depth images, in:The 4th International Symposium on Informatics and its Applications (ISIA 2020), IEEE, 2020, pp 1--6.

\bibitem{chuang2017image}
C.-H. Chuang, Y.-N. Chen, K.-C. Fan, Image feature point matching for indoor
  positioning, in: Proceedings of the International Conference on Image
  Processing, Computer Vision, and Pattern Recognition (IPCV), The Steering
  Committee of The World Congress in Computer Science, Computer~…, 2017, pp.
  139--140.

\bibitem{pujar2017combining}
K.~Pujar, S.~Chickerur, M.~S. Patil, Combining rgb and depth images for indoor
  scene classification using deep learning, in: 2017 IEEE International
  Conference on Computational Intelligence and Computing Research (ICCIC),
  IEEE, 2017, pp. 1--8.

\bibitem{williams2008image}
B.~Williams, M.~Cummins, J.~Neira, P.~Newman, I.~Reid, J.~Tard{\'o}s, An
  image-to-map loop closing method for monocular slam, in: 2008 IEEE/RSJ
  International Conference on Intelligent Robots and Systems, IEEE, 2008, pp.
  2053--2059.

\bibitem{williams2009comparison}
B.~Williams, M.~Cummins, J.~Neira, P.~Newman, I.~Reid, J.~Tard{\'o}s, A
  comparison of loop closing techniques in monocular slam, Robotics and
  Autonomous Systems 57~(12) (2009) 1188--1197.

\bibitem{hunter1979operations}
G.~M. Hunter, K.~Steiglitz, Operations on images using quad trees, IEEE
  Transactions on Pattern Analysis and Machine Intelligence~(2) (1979)
  145--153.

\bibitem{xing2018extracting}
Z.~Xing, Z.~Shi, Extracting multiple planar surfaces effectively and
  efficiently based on 3d depth sensors, IEEE Access 7 (2018) 7326--7336.

\bibitem{holzer2012adaptive}
S.~Holzer, R.~B. Rusu, M.~Dixon, S.~Gedikli, N.~Navab, Adaptive neighborhood
  selection for real-time surface normal estimation from organized point cloud
  data using integral images, in: 2012 IEEE/RSJ International Conference on
  Intelligent Robots and Systems, IEEE, 2012, pp. 2684--2689.

\bibitem{Holzer}
S.~Holzer, R.~B. Rusu, M.~Dixon, S.~Gedikli, N.~Navab, Adaptive neighborhood
  selection for real-time surface normal estimation from organized point cloud
  data using integral images, in: 2012 IEEE/RSJ International Conference on
  Intelligent Robots and Systems, IEEE, 2012, pp. 2684--2689.


\bibitem{fankhauser2015kinect}
P.~Fankhauser, M.~Bloesch, D.~Rodriguez, R.~Kaestner, M.~Hutter, R.~Siegwart,
  Kinect v2 for mobile robot navigation: Evaluation and modeling, in: 2015
  International Conference on Advanced Robotics (ICAR), IEEE, 2015, pp.
  388--394.

\bibitem{cummins2011appearance}
M.~Cummins, P.~Newman, Appearance-only slam at large scale with fab-map 2.0,
  The International Journal of Robotics Research 30~(9) (2011) 1100--1123.

\bibitem{nowakowski2020vision}
M.~Nowakowski, C.~Joly, S.~Dalibard, N.~Garcia, F.~Moutarde, Vision and wi-fi
  fusion in probabilistic appearance-based localization, The International
  Journal of Robotics Research (2020) 0278364920910485.

\bibitem{chen2019loop}
B.~Chen, D.~Yuan, C.~Liu, Q.~Wu, Loop closure detection based on multi-scale
  deep feature fusion, Applied Sciences 9~(6) (2019) 1120.

\bibitem{horst2017visual}
M.~Horst, R.~M{\"o}ller, Visual place recognition for autonomous mobile robots,
  Robotics 6~(2) (2017) 9.

\bibitem{glover2012openfabmap}
A.~Glover, W.~Maddern, M.~Warren, S.~Reid, M.~Milford, G.~Wyeth, Openfabmap: An
  open source toolbox for appearance-based loop closure detection, in: 2012
  IEEE International Conference on Robotics and Automation, IEEE, 2012, pp.
  4730--4735.
\bibitem{diffey2011overview}
Diffey, Brian L, An overview analysis of the time people spend outdoors, in 2011,
British Journal of Dermatology, 2011, pp.848--854.
\bibitem{mur2015orb}
Mur-Artal, Raul and Montiel, Jose Maria Martinez and Tardos, Juan D, ORB-SLAM: a versatile and accurate monocular SLAM system,
IEEE transactions on robotics journal, 2015, pp.1147--1163.


\bibitem{mur2017orb}
Mur-Artal, Raul and Tard{\'o}s, Juan D, Juan D, Orb-slam2: An open-source slam system for monocular, stereo, and rgb-d cameras,
IEEE transactions on robotics journal, 2017, pp.1255--1262.

\bibitem{zatout2021}
Zatout, C., Larabi, S. Semantic scene synthesis: application to assistive systems. Vis Comput (2021). https://doi.org/10.1007/s00371-021-02147-w

\bibitem{zatout2019ego}
Zatout, Chayma and Larabi, Slimane and Mendili, Ilyes and Ablam Edoh Barnabe, Soedji, Ego-semantic labeling of scene from depth image for visually impaired and blind people,
Proceedings of the IEEE/CVF International Conference on Computer Vision Workshops, 2019.

\bibitem{vatti1992generic}
B.~R. Vatti, A generic solution to polygon clipping, Communications of the ACM
  35~(7) (1992) 56--63.

\bibitem{hinze1999constructing}
Hinze, Ralf and others, Constructing red-black trees,
Proceedings of the Workshop on Algorithmic Aspects of Advanced Programming Languages, 1999, pp.89--99.

\bibitem{ulrich2000appearance}
Ulrich, Iwan and Nourbakhsh, Illah, Appearance-based place recognition for topological localization, Proceedings 2000 ICRA. Millennium Conference. IEEE International Conference on Robotics and Automation. Symposia Proceedings (Cat. No. 00CH37065), 2000, pp.1023--1029.
\bibitem{li2020precise}
Li, Ming and Chen, Ruizhi and Liao, Xuan and Guo, Bingxuan and Zhang, Weilong and Guo, Ge, A Precise Indoor Visual Positioning Approach Using a Built Image Feature Database and Single User Image from Smartphone Cameras, Multidisciplinary Digital Publishing Institute, 2020.

\bibitem{agarwal2010bundle}
Agarwal, Sameer and Snavely, Noah and Seitz, Steven M and Szeliski, Richard, Bundle adjustment in the large, European conference on computer vision, 2010, pp.29--42.

\bibitem{nowicki2017real}
Nowicki, Micha{\l} R and Wietrzykowski, Jan and Skrzypczy{\'n}ski, Piotr, Real-time visual place recognition for personal localization on a mobile device, 2017, pp.213--244.
\bibitem{chen2011city}
Chen, David M and Baatz, Georges and K{\"o}ser, Kevin and Tsai, Sam S and Vedantham, Ramakrishna and Pylv{\"a}n{\"a}inen, Timo and Roimela, Kimmo and Chen, Xin and Bach, Jeff and Pollefeys, Marc and others, City-scale landmark identification on mobile devices, 2011, pp.737--744.

\bibitem{jegou2008hamming}
Jegou, Herve and Douze, Matthijs and Schmid, Cordelia, Hamming embedding and weak geometric consistency for large scale image search, 2008, pp.304--317.
\bibitem{sturm12iros}
J. Sturm and N. Engelhard and F. Endres and W. Burgard and D. Cremers, A Benchmark for the Evaluation of RGB-D SLAM Systems, 2012.

\bibitem{larabi2009}
S. Larabi. Textual description of shapes,
Journal of Visual Communication and Image Representation,
Volume 20, Issue 8, 2009, Pages 563-584,
https://doi.org/10.1016/j.jvcir.2009.08.004.

\bibitem{Aouat2010}
S. Aouat and S. Larabi, "Indexing binary images using quad-tree decomposition," 2010 IEEE International Conference on Systems, Man and Cybernetics, 2010, pp. 3074-3080, doi: 10.1109/ICSMC.2010.5641701.

\end{thebibliography}

\end{document}